\documentclass[conference]{IEEEtran}

\IEEEoverridecommandlockouts
\usepackage{cite}
\usepackage{amsmath,amssymb,amsfonts}
\usepackage{algorithmic}
\usepackage{graphicx}
\usepackage{textcomp}
\usepackage{xcolor}
\usepackage{multirow}
\usepackage{booktabs} 
\usepackage{makecell}
\usepackage{pifont}
\def\BibTeX{{\rm B\kern-.05em{\sc i\kern-.025em b}\kern-.08em
    T\kern-.1667em\lower.7ex\hbox{E}\kern-.125emX}}
\begin{document}
\topmargin=0mm
\title{Intent-driven In-context Learning for Few-shot Dialogue State Tracking}

\author{
\IEEEauthorblockN{1\textsuperscript{st} Zihao Yi}
\IEEEauthorblockA{School of Intelligent Systems\\ Engineering \\
Sun Yat-sen University\\
Shenzhen, China \\
yizh6@mail2.sysu.edu.cn}
\and
\IEEEauthorblockN{2\textsuperscript{nd} Zhe Xu}
\IEEEauthorblockA{School of Intelligent Systems\\ Engineering \\
Sun Yat-sen University\\
Shenzhen, China \\
xuzh226@mail2.sysu.edu.cn
}
\and
\IEEEauthorblockN{3\textsuperscript{rd} Ying Shen$^{\ast}$\thanks{*Ying Shen is the corresponding author.}}
\IEEEauthorblockA{School of Intelligent Systems\\ Engineering \\
Sun Yat-sen University\\
Shenzhen, China \\
sheny76@mail.sysu.edu.cn}
}

\maketitle

\begin{abstract}
Dialogue state tracking (DST) plays an essential role in task-oriented dialogue systems. However, user's input may contain implicit information, posing significant challenges for DST tasks. Additionally, DST data includes complex information, which not only contains a large amount of noise unrelated to the current turn, but also makes constructing DST datasets expensive. To address these challenges, we introduce Intent-driven In-context Learning for Few-shot DST (IDIC-DST). By extracting user's intent,  we propose an Intent-driven Dialogue Information Augmentation module to augment the dialogue information, which can track dialogue states more effectively. Moreover, we mask noisy information from DST data and rewrite user's input in the Intent-driven Examples Retrieval module, where we retrieve similar examples. We then utilize a pre-trained large language model to update the dialogue state using the augmented dialogue information and examples. Experimental results demonstrate that IDIC-DST achieves state-of-the-art performance in few-shot settings on MultiWOZ 2.1 and MultiWOZ 2.4 datasets.
\end{abstract}

\begin{IEEEkeywords}
dialogue state tracking, few-shot, large language model, in-context learning, intent extraction
\end{IEEEkeywords}

\section{INTRODUCTION}
\label{sec:intro}

Task-oriented dialogue (TOD) has become increasingly integral to our daily lives, assisting users in tasks such as booking flights and planning trips. A typical TOD system consists of four modules \cite{c1}: (1) Natural Language Understanding (NLU); (2) Dialogue State Tracking (DST); (3) Policy Learning; and (4) Natural Language Generation.
DST is the central component of TOD systems, where it tracks the user's intent using slot-value pairs to represent the dialogue state.

Construction DST datasets is both expensive and time-consuming \cite{survey}. But most DST methods\cite{DSCL-DST,LAL-DST,SSNET} require large amounts of data to achieve satisfactory performance. Although some methods \cite{STN4DST,LUNA} have enhanced the model's generalization performance in unknown domains, these models still heavily rely on DST datasets. Therefore, it is crucial to research effective few-shot DST methods to mitigate the expenses associated with constructing DST datasets.

Recently, more and more researchers focus on large language models (LLMs) since they demonstrate superior performance in few-shot scenario \cite{LLM}. Kulkarni et al.\cite{SYNTHDST} accomplishes the DST task by automatically generating context to guide the LLMs. Yang et al. \cite{DPL} utilizes the LLMs to independently generate potential slots and values. Su et al. \cite{CoFunDST} train LLMs on QA datasets and directly use candidate choices of slot-values as knowledge for zero-shot dialogue-state generation. Lee et al. \cite{CONVERSE} leverage LLMs to summarize dialogues for completing DST tasks.

\begin{table}[t]
    \centering
    \caption{Task-oriented Dialogue Universal Data Format. \textcolor{red}{Red} means implicit content. \textcolor{blue}{Blue} means content which may contain noise. $B_t$ refers to dialogue state and $I_t$ refers to user's intent.}
    \begin{tabular}{|c|c|}
        \hline
         turn& 2 \\
         \hline
         domain&[``attraction"] \\
         \hline
         \textcolor{blue}{context}& \makecell{usr: [
                ``i am looking for attractions in Hyderabad ."
            ]\\
            sys: [
                ``we have many, which area do you prefer?'']\\
            \textcolor{red}{\textbf{usr: [
                ``I want 1 in the south area."}
            ]}}\\
            \hline
         \textcolor{blue}{$B_t$}&\{
            ``attraction-city":``Hyderabad", ``attraction-area":``south"
        \} \\
         \hline
         $I_t$&[inform]\{
            ``attraction-area":``south"
        \} \\
                 \hline
         \textcolor{blue}{other}&...... \\
         
         \hline
    \end{tabular}
    \label{tab:data_item}
    \vspace{-\baselineskip}
\end{table}

However, due to the presence of bias and noise in the pre-training data, LLMs may struggle to perform well in unseen domains \cite{Hallucination}. Additionally, as shown in table \ref{tab:data_item}, DST data includes dialogue history and a significant amount of other information, which can contain substantial noise. To improve the performance of LLM-based few-shot DST, we analyze many few-shot DST methods and identified the following limitations in few-shot DST methods:
\begin{enumerate}
    \item \textbf{Implicit information extraction}: As shown in table \ref{tab:data_item}, user's input may contain implicit information. However, most few-shot DST methods simply take dialogue context and historical dialogue states as input, which makes it challenging for these methods to accurately extract 
    user's implicit intent.  This difficulty arises because these methods heavily rely on clear dialogue context, making adaptation to deceptive dialogues challenging. 
    \item \textbf{In-context examples retrieval}: DST data comprises various types of information, such as dialogue turns, and historical dialogue states, which may contain significant noise.  However, existing in-context learning-based DST methods \cite{ICDST,REFPY} simply retrieve examples based on dialogue context and historical dialogue states. On the one hand, dialogue context includes the entire dialogue history, which may introduce noisy information. On the other hand, user's input may contain implicit information, which may result in inappropriate examples.

\end{enumerate}

To address these challenges, we propose Intent-driven In-context Learning for Few-shot DST (IDIC-DST) that enhances few-shot DST performance in two key ways: (1) by extracting the user’s current dialogue intent and appending it to the dialogue information, thereby improving its ability to handle implicit information; and (2) by masking irrelevant dialogue information and rewriting user's input to increase the emphasis on key information, thus reducing errors caused by noise when retrieving in-context examples. Experimental results demonstrate that IDIC-DST outperforms existing few-shot DST methods on MultiWOZ 2.1 and 2.4 datasets, enabling more efficient and accurate completion of TOD.

\section{METHOD}

\begin{figure}
    \centering
    \includegraphics[width=1\linewidth]{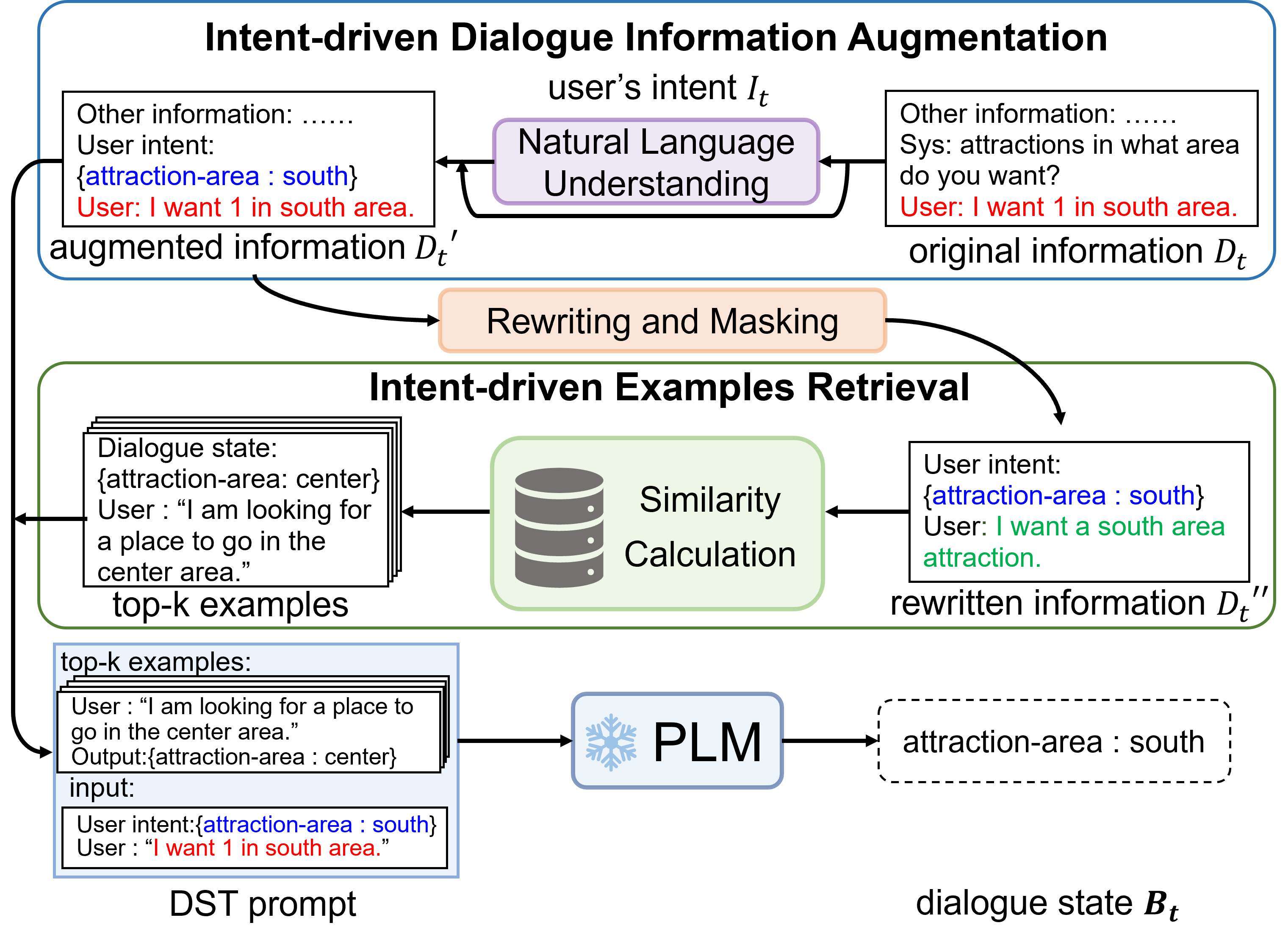}
    \caption{Overview of IDIC-DST method. The Intent-driven Dialogue Information Augmentation module extracts  \textcolor{blue}{user's intent} from \textcolor{red}{implicit input} to augment the dialogue information, while the Intent-driven Examples Retrieval module  \textcolor{green}{rewrites user's input} to better retrieve $k$ similar dialogue examples. The augmented dialogue information ${D_t}^{'}$ is then combined with the in-context examples to generate DST prompts, which are used as input to a pre-trained LLM to update the dialogue state $B_t$.}
    \label{fig:1}
    \vspace{-\baselineskip}
\end{figure}
DST task predicts the dialogue state at turn \(t\), denoted as \(B_t = \{(s_1, v_1), (s_2, v_2), \dots, (s_n, v_n)\}\), where \((s_n,v_n)\) is a slot-value pair. For example,  \{(\text{train-arriveat}, \text{10:00}), (\text{train-leaveat}, \text{7:00})\}. The input of DST task is the dialogue information, denoted as \(D_t = \{U_t, H_{t-1}, B_{t-1}, O_t\}\), where \(U_t\) represents the user's input, \(H_{t-1}\) represents the dialogue history and \(O_t\) includes other information such as dialogue turn and dialogue domain. DST task can be formalized as:

\begin{equation}
    B_t = \text{DST}(U_t, H_{t-1}, B_{t-1},O_t).
\end{equation}

Figure \ref{fig:1} presents the architecture of the IDIC-DST framework, which is composed of two main modules: the Intent-driven Dialogue Information Augmentation module and the Intent-driven Examples Retrieval module.

In the Intent-driven Dialogue Information Augmentation module, we first extract the user's intent \( I_t \) from \( U_t \) using a fine-tuned T5 model \cite{T5}. \( I_t \) is then incorporated into \( D_t \), resulting in the augmented dialogue information \( D_t' \).

Following this, in the Intent-driven Examples Retrieval module, specific tokens in \( D_t' \) are masked, generating the masked dialogue information \( D_t'' \). We then calculate the similarity between \( D_t'' \) and other dialogues within the dataset. The top \( k \) examples with the highest similarity scores are selected. These selected examples are subsequently concatenated with \( D_t' \) and fed into a pre-trained LLM, which is then used to update \( B_t \).

\subsection{Intent-driven Dialogue Information Augmentation}
NLU focuses on identifying user's intents and extracting dialogue domains from dialogues, enabling computers to understand and interpret natural language. Therefore, we enhance the performance of LLMs on DST tasks through NLU models.

Specifically, we fine-tune a T5 model as a NLU model to extract $I_t$ from $U_t$, which is then concatenated with $D_t$, producing the augmented dialogue information $D_t'$. The computation of $D_t'$ can be formalized as:

\begin{equation}
    D_t' = D_t \oplus \text{NLU}(H_{t-1}, B_{t-1}, U_t, O_t).
\end{equation}

To extract the user's intent, we employ a T5-small model, which is trained in a few-shot setting using TOD dataset. Specifically, we utilize dialogue context as input, with the user's intent $I_t$ as the output. This training process results in a T5 model that is specifically fine-tuned for NLU tasks.

\subsection{Intent-driven Examples Retrieval}
We test the impact of different information from $D_t$ on retrieval results and we find that: 
\begin{enumerate}
    \item In-context examples retrieved based on $I_t$ provide the most effective guidance for the model. This is likely due to examples that closely match the current user's intent offer more relevant information. The dialogue state $B_{t-1}$, which comprises slot-value pairs from previous turns, may introduce irrelevant information, diminishing the effectiveness of guidance. Thus, in-context examples retrieved solely based on the current turn's slot-values $I_t$ outperforms those retrieved based on other information.
    \item In-context examples retrieved based on dialogue history show the lowest relevance to current turn. This is likely because the dialogue history contains the complete interaction between the user and the system, which includes a substantial amount of information irrelevant to the current user's intent. Therefore,  the dialogue history may be unnecessary for Intent-driven Examples Retrieval.
\end{enumerate}

Based on above findings, we mask dialogue history, $B_t$ and $B_{t-1}$ before retrieval to obtain the rewritten information $D_t^{''}$:

\begin{table*}[ht]
    \centering
    
    \tabcolsep=0.03\linewidth
    \caption{Joint goal accuracy (\%) on MultiWOZ 2.1 and MultiWOZ 2.4 in 1\% few-shot setting.}
    \begin{tabular}{l|c|c|c}
    \toprule
        Model& \makecell{Parameter Size} &  \makecell{1\% MultiWOZ 2.1} &  \makecell{1\% MultiWOZ 2.4} \\
        \hline 
        TRADE (2019 ACL)\cite{TRADE}& 
\multirow{4}{*}{$<1$B}& 12.58  &  -  \\
        SGPDST (2019 EMNLP)\cite{SGPDST}& & 32.11  &  -  \\
        DS2 (2022 ACL)\cite{DS2}& & 33.76  &  36.76  \\
        FaS-DST (2024 ICASSP)\cite{FaS-DST}&&36.27&-\\
        \hline 
        IC-DST GPT-Neo (2022 EMNLP)\cite{ICDST}&\multirow{4}{*}{$1\sim10$B} & 16.70  &  17.36  \\
        SM2-11B (2023 ACL)\cite{SM2}&  &38.06  &  37.59  \\
        LDST (2023 EMNLP)\cite{LDST}&  &-  &  46.77  \\
        \textbf{IDIC-DST (ours)}&  &\textbf{43.11}  &  \textbf{52.66}  \\
        \hline 
        SM2-11B (2023 ACL)\cite{SM2}&\multirow{1}{*}{$10\sim100$B} & 38.36  &  40.03  \\
        \hline 
         IC-DST Codex (2022 EMNLP)\cite{ICDST}&\multirow{1}{*}{$>100$B} & 43.13  &  48.35  \\

    \bottomrule
    \end{tabular}
    
    \label{tab:对比1}
    \vspace{-\baselineskip}
\end{table*}

\begin{equation}
\begin{aligned}
D_t^{''} &= \text{Mask}(D_t') \\
&= \{ \text{turn}, \text{domain}, [\text{MASK}], [\text{MASK}],I_t, [\text{MASK}] \}.
\end{aligned}
\end{equation}

Then, we perform a simple rewrite of the user's input based on $I_t$ to ensure that the user's input includes explicit dialogue information, which will also be used for retrieving examples. For example, given $I_t$ : \{ ``train-destination":``Hyderabad"\}, the user's input is rewritten as ``I want a train to Hyderabad.".

To enhance the retrieval performance in DST tasks, we fine-tune a Sentence-BERT (SBERT) model in a few-shot setting, employing it as a retriever. Consider two sets of state changes, $c_a = \{(s_{a}^1, v_{a}^1), \dots, (s_{a}^m, v_{a}^m)\}$ and $c_b = \{(s_{b}^1, v_{b}^1), \dots, (s_{b}^n, v_{b}^n)\}$, where each element $(s, v)$ represents a slot-value pair. 
The slot similarity can be formalized as:
\begin{equation}
    F_{\text{slot}} = F(\{s_{a}^1, \dots, s_{a}^m\}, \{s_{b}^1, \dots, s_{b}^n\}),
\end{equation}
where $F(set1, set2)$ is the average $F1$ score between $set_1$ and $set_2$. Similarly, the slot-value similarity can be defined as:
\begin{equation}
\begin{aligned}
    F_{\text{slot-value}} = F(&\{(s_{a}^1, v_{a}^1), \dots, (s_{a}^m, v_{a}^m)\},\\&\{(s_{b}^1, v_{b}^1), \dots, (s_{b}^n, v_{b}^n)\}).
\end{aligned}
\end{equation}

The similarity score $S(c_a, c_b)$ can be calculated by:
\begin{equation}
\begin{aligned}
S(c_a, c_b) = \frac{1}{2}(F_{\text{slot}} + F_{\text{slot-value}}).
\end{aligned}
\end{equation}

The retriever is then fine-tuned using contrastive loss to ensure high similarity scores between similar sets of state changes. The retriever is subsequently used to calculate the similarity between examples and the current dialogue, and the top $k$ scoring examples are selected to construct DST prompts.

\subsection{Dialogue State Updating}

Numerous prior studies \cite{ICDST,REFPY} have approached DST task by framing it as a code generation problem, where dialogue states are dynamically updated by generating code. Building on these approaches, we propose modeling the DST task as a text-to-SQL generation task, leveraging the SQL language to dynamically update dialogue states. 

Our method follows a specific set of rules:
\begin{enumerate}
    \item Each table in the SQL schema represents a domain, while each column in the table corresponds to a slot.
    \item The \texttt{WHERE} clause captures all changes in the dialogue state relevant to the current conversation.
    \item If the current dialogue involves \(n\) domains, each domain is represented by a unique alias \(d_1, \dots, d_n\).
\end{enumerate}

The input prompt \(P_t\) for the LLM consists of a predefined SQL command, which initializes tables to define the dialogue domains and slot values. This fixed SQL command is then augmented with the top-\textit{k} retrieved in-context examples and the augmented dialogue information \(D_t''\). The pre-trained LLM processes this DST prompt to generate a SQL query, which reflects the updates required for the current dialogue state. Subsequently, the specific changes in the dialogue state \(S_t\) are extracted from the generated SQL query. The process of updating \(B_t\) can be formalized as follows:

\begin{equation}
\begin{aligned}
&SQL_t = PLM(P_t),\\
&S_t = sql(SQL_t, B_{t-1}),
\end{aligned}
\end{equation}
where \(sql( \cdot)\) is the SQL query extraction function.

\section{EXPERIMENTS}

\subsection{Datasets \& Metrics}
We employ the MultiWOZ 2.1 and 2.4 \cite{2.1,2.4}  datasets to experimentally evaluate our proposed IDIC-DST method. The MultiWOZ dataset is a dialogue dataset widely used in research on task-oriented multi-turn dialogues within the field of natural language processing. It contains 10,438 dialogues spanning 7 different domains, including restaurants, hotels, attractions, taxis, and transportation. Each dialogue covers 1 to 5 domains, with an average of 13.7 turns per dialogue, exhibiting substantial variation in length and complexity.

We use the Joint Goal Accuracy (JGA) \cite{JGA} as the evaluation metric. For each turn in a dialogue, a dialogue state is considered correct only if it exactly matches the ground truth, meaning all slots and values are correctly predicted. 

Additionally, we evaluate IDIC-DST in practical dialogue tasks using four other metrics: \textit{booking rate},  representing the percentage of dialogues that result in successful bookings; \textit{F1 score}, which balances precision and recall to assess dialogue state accuracy; \textit{completion rate}, indicating the proportion of dialogues that finish within a restricted turn; and \textit{average turn}, counting the average turn needed to complete dialogues.

\begin{table*}[ht]
    \centering
    \tabcolsep=0.02\linewidth
    \caption{Performance of different DST methods in task-oriented multi-turn dialogue on MultiWOZ 2.1.}
    \scalebox{1.1}{ %此处放置命令
    \begin{tabular}{l|cccc}
    \toprule
        Model & Completion Rate$ \uparrow$ & Booking Rate$ \uparrow$ & F1$ \uparrow$ & Average Turn$ \downarrow$\\
        \hline 
        Trippy (2020 SIGDIAL)\cite{trippy}  & 0.42 & 0.48 &  0.51 & 17\\
        IC-DST GPT-Neo (2022 EMNLP)\cite{ICDST} & 0.5 & 0.41 & 0.56 & 19.2\\
        T5dst (2023 EMNLP)\cite{convlab3}  & 0.71 & 0.51 & 0.52 & 17.5\\
        \hline 
        IDIC-DST (ours) & \textbf{0.81} & \textbf{0.82} & \textbf{0.64} & \textbf{16.4}\\
        \bottomrule
    \end{tabular}
    }
    \label{table:evaluation}
    \vspace{-\baselineskip}
\end{table*}

\subsection{Implementation Details}
To validate the performance of IDIC-DST, we extracted 1\% of the MultiWOZ dataset for training the NLU model and the retriever, which is also used as the retrieval sample pool.

The NLU model is initialized using the T5-small model \cite{T5}, with a batch size of 128, a learning rate of $10^{-3}$, 10 epochs, and the Adafactor optimizer. The retriever is initialized using the SBERT all-mpnet-v2 model \cite{SBERT}, with a batch size of 24, a learning rate of $2 \times 10^{-5}$, 15 epochs, and the Adam optimizer. For the dialogue state update phase, codeLlama 7B\cite{CLLAMA} is selected as the generation model.
\begin{table}[t]
    \centering
    \tabcolsep=0.04\linewidth
    \caption{Ablation study.``IDA" represents adding Intent-driven Dialogue Information Augmentation module. ``IER" represents adding Intent-driven Examples Retrieval module.}
\scalebox{1.1}{ %此处放置命令
    \begin{tabular}{l|cccc}
    \toprule

         Model&    1\%&  5\%&  10\%&100\% \\
         \hline 
        w/o DCA,IER &  21.59 & 32.04 & 32.98 &37.36 \\
         \hline 
         w/ DCA & 33.14 & 39.59 & 41.16 & 41.82 \\
         \hline 
         w/ DCA,IER &  \textbf{43.39} & \textbf{44.25} & \textbf{43.96} & \textbf{44.93} \\
    \bottomrule
    \end{tabular}
    }
    \label{tab:evaluation}
    \vspace{-\baselineskip}
\end{table}
\subsection{Main Results}

Table \ref{tab:对比1} shows the results of IDIC-DST compared with baseline methods. It is evident that our model outperforms all baseline models with parameter sizes less than 100B. When compared to the SOTA DST method SM2-11B, which has over 10 B parameters, IDIC-DST demonstrates a \textbf{4.75\%} improvement in JGA on MultiWOZ 2.1 and an even more substantial \textbf{12.63\%} advantage in JGA on MultiWOZ 2.4.
Even when compared with the IC-DST Codex, which utilizes  Codex-Davinic model with 175B parameters, IDIC-DST achieves comparable performance on MultiWOZ 2.1 dataset and even demonstrates a \textbf{4.31\%} improvement on MultiWOZ 2.4.

To validate the performance of IDIC-DST in TOD tasks, we implemented a complete TOD system. The NLU module utilizes T5 NLU model \cite{convlab3}, the policy learning module adopts a dynamic dialogue policy transformer \cite{DDPT} model, and the natural language understanding module employs a template-based generation method. 

As shown in Table \ref{table:evaluation}, the results indicate that the TOD system based on IDIC-DST achieves superior performance in terms of all evaluation metrics. This demonstrates that IDIC-DST can more accurately extract user's intent, improving the accuracy and efficiency of TOD.

\subsection{Ablation Study}

As shown in table \ref{tab:evaluation}, we conducted ablation experiments based on IC-DST on MultiWOZ 2.4 dataset.

\subsubsection{Intent-driven Dialogue Information Augmentation}
The introduction of Intent-driven Dialogue Information Augmentation module significantly improved the JGA score of IC-DST on MultiWOZ 2.4, which achieves an impressive \textbf{11.55\%} increase in 1\% few-shot setting.

\subsubsection{Intent-driven Examples Retrieval}
After integrating the Intent-driven Examples Retrieval module and the Intent-driven Dialogue Information Augmentation module, IC-DST demonstrates substantial improvement on the MultiWOZ 2.4 dataset, particularly achieving a \textbf{21.8\%} improvement in the 1\% few-shot setting. This result fully validates the effectiveness of our proposed Intent-driven Examples Retrieval module.
\begin{table}
    \centering
    \caption{One case to illustrate the effectiveness of IDIC-DST. \textcolor{orange}{Orange} means sentence containing implicit information. \textcolor{red}{Red} means wrong massages. \textcolor{blue}{Blue} means  explicit or relevant content.}
    \begin{tabular}{|p{8cm}|}
    \hline
         \textbf{Dialogue:} [User] i am looking for attractions to go to in town . [System] we have 4 of in the centre and 1 in the south. any area preference? [User]\textcolor{orange}{can you give me some info on the 1 in the south.}\\
         \textbf{Golden output:}(
        attraction-area:south
        )\\
        \hline
        \textbf{IC-DST retrieval context}: [CONTEXT] \{attraction type: theatre\}, [System] we have 4 of in the centre and 1 in the south. any area preference? [User]\textcolor{orange}{can you give me some info on the 1 in the south.}
\\
        \textbf{IC-DST Example}:
[user] what is the postcode ?
 \textcolor{red}{\textbf{Dialogue State: None}}
\\
\textbf{IC-DST output}: \textcolor{red}{\textbf{None}} (\textcolor{red}{\ding{55}})\\
\hline
\textbf{IDIC-DST retrieval context}: [CONTEXT] \{ \textcolor{blue}{\textbf{attraction area: south}} \} [SYS] [USER] i an looking for a south area attraction. [DOMAIN] attraction
\\
\textbf{IDIC-DST Example}: 
[user] i am also looking for place -s to go in the centre .
\textcolor{blue}{\textbf{Dialogue State: (attraction-area:center)}}\\
\textbf{IDIC-DST referance output}: \textcolor{blue}{\textbf{ (attraction-area:south)}} 
        \\
\textbf{IDIC-DST final output}: \textcolor{blue}{\textbf{ (attraction-area:south)}} (\textcolor{green}{\checkmark})
        \\
    \hline
    \end{tabular}
    \label{tab:case}
    \vspace{-\baselineskip}
\end{table}
\subsection{Case Study}
As shown in table \ref{tab:case},  we conducted a case study to evaluate the performance of IDIC-DST and the SOTA method IC-DST in specific dialogue scenarios. The IC-DST method retrieves $B_{t-1}$ and dialogue history. However, it encounters difficulties in retrieval when the user's input lacks explicit domain information. In contrast, IDIC-DST not only retrieves $I_t$ and rewrites the user's input based on it, thereby \textbf{retrieving more suitable in-context examples}, but also provide $I_t$ as a reference output to the LLM, enabling it to \textbf{perform well even when the user's input is implicit}. This approach effectively guides LLMs to generate correct outputs.

\section{CONCLUSION}

In this paper, we introduce Intent-driven In-context Learning for Few-shot DST (IDIC-DST) to address the challenges posed by implicit dialogue information and data complexity in TOD systems. By extracting user's intent, IDIC-DST enriches dialogue information while effectively masking noisy information, thereby improving the alignment between retrieved in-context examples and the current dialogue context. 
Experimental results on the MultiWOZ 2.1 and MultiWOZ 2.4 datasets demonstrate that IDIC-DST achieves SOTA performance in few-shot settings, highlighting its effectiveness in enhancing DST accuracy under challenging conditions.

\end{document}